\pdfoutput=1

\documentclass[11pt]{article}

\usepackage[]{EMNLP2023}

\usepackage{times}
\usepackage{latexsym}

\usepackage[T1]{fontenc}

\usepackage[utf8]{inputenc}

\usepackage{microtype}

\usepackage{inconsolata}

\pdfoutput=1
\usepackage{multirow}
\usepackage{colortbl}

\usepackage{times}
\usepackage{latexsym}
\usepackage{xspace}

\usepackage[T1]{fontenc}

\usepackage[utf8]{inputenc}

\usepackage{microtype}

\usepackage{amsmath}
\usepackage{amsfonts}
\usepackage{graphicx,subfigure}
\usepackage{ctable}
\usepackage{comment}
\usepackage{multicol}
\usepackage{multirow}
\usepackage{epsfig}
\usepackage{pifont}
\usepackage{dsfont}
\usepackage{makecell}
\usepackage{adjustbox}
\usepackage{txfonts}

\usepackage[boxed, ruled, vlined, linesnumbered]{algorithm2e}
\usepackage{amstext}
\usepackage{xspace}
\usepackage{xcolor}
\usepackage{pict2e}

\usepackage{amssymb}
\usepackage{pifont}

\newcommand*{\img}[1]{%
    \raisebox{-.15\baselineskip}{%
        \includegraphics[
        height=0.9\baselineskip,
        width=\baselineskip,
        keepaspectratio,
        ]{#1}%
    }%
}

\title{\system \hspace*{-0.08in}
\img{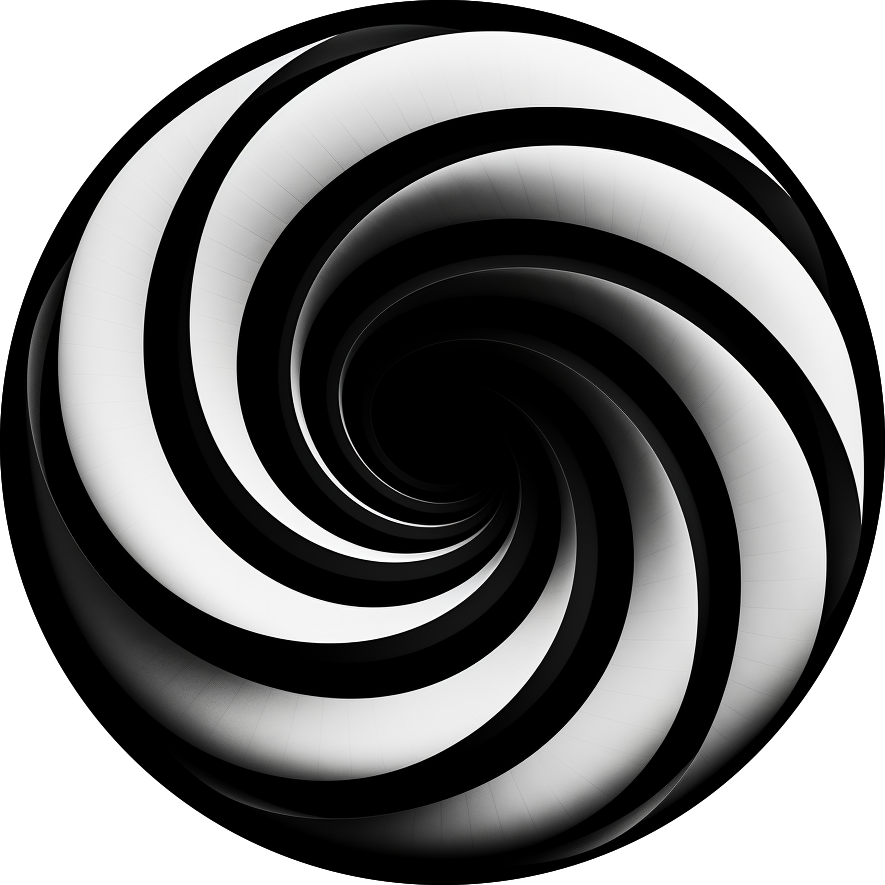}: A Collaborative Platform for Tool-Augmented LLMs }

\author{
{\normalsize 
Binfeng Xu$^*$ ~ Xukun Liu ~ Zeyu Han ~ Hua Shen ~ Yuhan Li ~ Zhiyuan Peng ~ Murong Yue\vspace{0.2mm} 
}\\ 
{\normalsize\bf Ziyu Yao ~ Yuchen Liu ~ Dongkuan Xu
\vspace{0.2mm}
}\\
}

\author{
{\normalsize 
Binfeng Xu, ~ Xukun Liu, ~ Hua Shen, ~ Zeyu Han, ~ Yuhan Li, ~ Murong Yue\vspace{0.2mm}, ~ Zhiyuan Peng,
}\\ 
{\normalsize\bf Yuchen Liu, ~ Ziyu Yao, ~ Dongkuan Xu
\vspace{0.2mm}
}\\
{\normalsize \url{https://github.com/Gentopia-AI}}
}

\begin{document}

\newcommand{\bill}[1]{{\small\textcolor{blue}{\bf [#1 --Bill]}}}
\newcommand{\yuan}[1]{{\small\textcolor{cyan}{\bf [#1 --Yuan]}}}
\newcommand{\hua}[1]{{\small\textcolor{orange}{\bf [#1 --Hua]}}}
\newcommand{\ziyu}[1]{{\small\textcolor{violet}{\bf [#1 --Ziyu]}}}
\newcommand{\dk}[1]{{\small\textcolor{olive}{\bf [#1 --DK]}}}
\newcommand{\graham}[1]{{\small\textcolor{brown}{\bf [#1 --Graham]}}}
\newcommand{\lyh}[1]{{\small\textcolor{red}{\bf [#1 --LYH]}}}

\newcommand{\cmark}{\ding{51}}%
\newcommand{\xmark}{\ding{55}}%

\newcommand{\yes}{\textcolor[HTML]{589C30}{\cmark}}%

\newcommand{\no}{\textcolor[HTML]{ED6E72}{\xmark}}%

\newcommand{\system}{\texttt{Gentopia.AI}\xspace}

\newcommand{\gentopia}{\texttt{Gentopia}\xspace}
\newcommand{\gentbench}{\texttt{GentBench}\xspace}
\newcommand{\gentpool}{\texttt{GentPool}\xspace}

\maketitle

\begin{abstract}

Augmented Language Models (ALMs) empower large language models with the ability to use tools, transforming them into intelligent agents for real-world interactions. However, most existing frameworks for ALMs, to varying degrees, are deficient in the following critical features: flexible customization, collaborative democratization, and holistic evaluation. We present \gentopia, an ALM framework enabling flexible customization of agents through simple configurations, seamlessly integrating various language models, task formats, prompting modules, and plugins into a unified paradigm. Furthermore, we establish \gentpool, a public platform enabling the registration and sharing of user-customized agents. Agents registered in \gentpool are composable such that they can be assembled together for agent collaboration, advancing the democratization of artificial intelligence. 
To ensure high-quality agents, \gentbench, an integral component of \gentpool, is designed to thoroughly evaluate user-customized agents across diverse aspects such as safety, robustness, efficiency, etc. We release \gentopia on Github\footnote{\url{https://github.com/Gentopia-AI/Gentopia}. All mentioned works are under MIT license. Check our demo at
\url{https://www.youtube.com/watch?v=7dZ3ZvsI7sw} and homepage at \url{https://gentopia-ai.github.io/Gentopia-AI-Homepage/}.} and will continuously move forward. 

\end{abstract}

\section{Introduction}
\label{sec:introduction}

There is a burgeoning trend in research towards augmenting large language models (LLMs) with external tools, enabling them to access up-to-date databases~\cite{jiang2023structgpt,pan2023unifying}, perform arithmetic operations~\cite{imani2023mathprompter}, navigate websites~\cite{gur2023real}, develop software~\cite{wu2023metagpt}, etc. This integration of tools marks a departure from traditional language modeling, heralding a new era of intelligent agents capable of interacting with the real world.

\begin{figure*}[!t]
  \includegraphics[width=\textwidth]{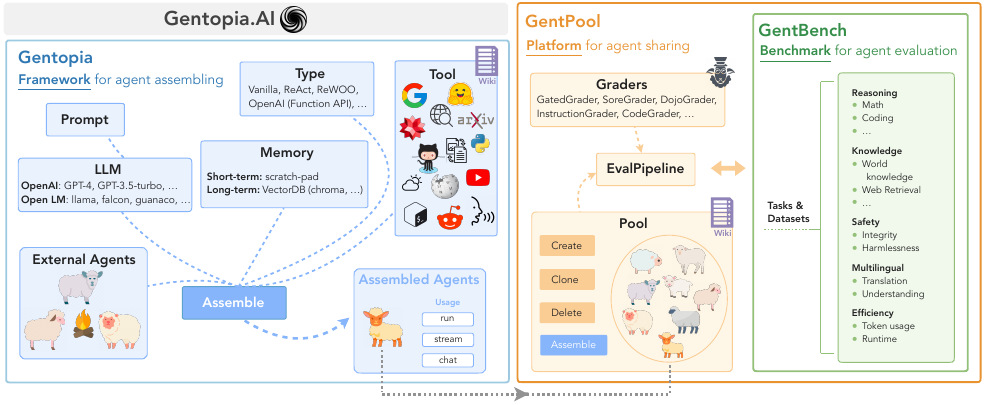}
  \caption{An overview of \system, encapsulating following pivotal components: 1) \textbf{\gentopia}: a framework principally designed to assemble an agent instance from a YAML configuration file, composed of multiple pre-built agent components such as the LLM, tools, memory, and external agents; 2) \textbf{\gentpool}: a platform engineered to facilitate the registration and sharing of specialized agents, seamlessly integrating \textbf{\gentbench}, an ALM benchmark devised specifically for the comprehensive performance evaluation of agents.}
  \vspace{-1em}
  \label{fig:overview}
\end{figure*}

Several projects and frameworks have been proposed to build tool-Augmented Language Models (ALMs), or "agents", including AutoGPT~\cite{richards2023auto}, SuperAGI~\cite{superagi}, HuggingGPT~\cite{shen2023hugginggpt}, GPT-Engineer~\cite{osika2023gptengineer}, LangChain~\cite{langchain2023}, Semantic Kernel~\cite{callegari2023semantic}, and MiniChain~\cite{srush2023minichain}. 
Each of these methods is deficient, to varying degrees, in the following critical features.







\begin{itemize}
    \item \textbf{Adaptive Customization}: Many are designed for a single set of tasks without extensive support in customization, or they involve redundant and boilerplate implementation that unnecessarily complicates agent tuning.

    \item \textbf{Tool-augmented NLP Benchmark}: A user-customized agent, before registration, is expected to go through a thorough evaluation to ensure its quality. However, there is a lack of comprehensive benchmarks designed for agent evaluation in the aspects of efficiency, safety, robustness, etc.

    \item \textbf{Democratization}: A platform where user-customized agents can be registered and shared is missing. This hinders the interaction and collaboration of various user-customized agents. Collaborative growth is a critical point toward safe and powerful intelligence.

\end{itemize}
This paper proposes \gentopia, a lightweight and extensible framework for the research on ALMs. \gentopia allows practitioners to customize an agent with a single configuration file, greatly simplifying the process of building, tuning, sharing, and evaluating agents. Various language models, task formats, prompting modules, and plugins are integrated into a unified paradigm, without loss of flexibility for agent customization. In addition, we believe the collaboration between user-customized agents can contribute to the democratization of AI. Hence, \gentpool, a platform for agent registration and sharing is established. Agents registered in \gentpool can be hierarchically assembled together, enabling the collaboration of multiple agents. \gentpool is accompanied by a unique benchmark, \gentbench, that can probe customized agents with a holistic evaluation in terms of safety, robustness, efficiency, multilingual capabilities, etc. Notably, it is flexible for users to customize the evaluation by configuration.






\section{Background}

\label{sec:literature}

A variety of agent projects have been proposed, targeting an array of diverse tasks, including automated web navigation~\cite{gur2023real}, database management~\cite{jiang2023structgpt}, automated game playing~\cite{wang2023voyager}, collaborative software development~\cite{wu2023metagpt}, etc. Meanwhile, researchers are enthusiastically developing generalist agents that can perform well for multiple tasks. AutoGPT~\cite{richards2023auto} stands for the first experimental open-source application for fully automatic AI, with the ultimate goal of ``autonomously achieving whatever goal users set". SuperAGI~\cite{superagi} provides a more user-friendly interface, improved memory management, optimized token usage, and looping detection heuristics. HuggingGPT~\cite{shen2023hugginggpt} expands the potential of artificial intelligence by linking to extensive AI models hosted on HuggingFace, thereby supporting a range of AI tasks in diverse domains and modalities, including language, vision, and speech.

However, given the unique requirements and customization that each specific domain demands, tools and prompting paradigms developed for a particular task may prove irrelevant or ineffective for others. This poses a significant challenge to the development of a single, all-encompassing agent that performs efficiently across all tasks. Consequently, there is a rising need for the collaboration of multiple specialized agents. For example, MetaGPT~\cite{wu2023metagpt} models the entire process of software development with carefully orchestrated standard operating procedures (SOPs) to generate longer program codes for game development. In our work, \gentopia provides smooth support for the composition of agents, which is handy for agent collaboration in different scenarios.

\section{Design and Implementation}
\label{sec:system}
\gentopia aims to provide easy assembly, sharing, and interaction of task-specialized agents. A single step to improve agent capability and efficiency gives plural contributions to interacted agents,  fostering collective growth toward greater intelligence.
%


\begin{figure*}[!t]
    \centering
  \includegraphics[width=.92\textwidth]{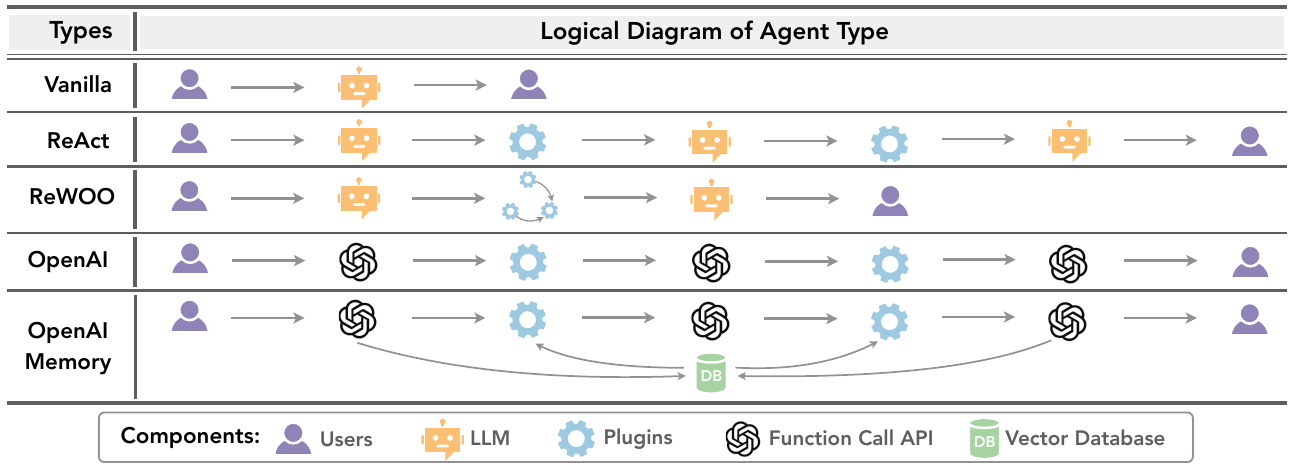}
  \caption{\gentopia agent types, categorized according to the interaction paradigms between agents and plugins.}
  \label{fig:agent-types}
  \vspace{-.5em}
\end{figure*}

\subsection{Rationale}
The impetus of \gentopia is rooted in the aspiration to construct capable and deployable AI assistants. A pertinent question that arises in this context is whether there is a necessity for a massive and expensive model like 175B GPT-4 to perform relatively simple tasks such as summarizing a web search. Recent studies like TinyStories~\cite{eldan2023tinystories}, Specializing Reasoning~\cite{fu2023specializing}, Let’s Verify Step by Step~\cite{lightman2023let}, and ReWOO~\cite{xu2023rewoo}, direct our attention towards an intuitive yet undervalued observation -- LLMs exhibit enhanced capabilities when a context/distribution shift is created, specifically narrowed towards certain target tasks.

However, there is no silver bullet for agent specialization. Various strategies can be employed depending on target tasks. For instance, prompting "Let's think step by step" in context leads to more accurate math reasoning~\cite{kojima2022large}. Providing few-shot examples could guide an ideal execution workflow. Instruction tuning allows an LLM to excel on fine-tuned datasets or tasks~\cite{wei2021finetuned}. Tweaking the agent type from ReAct~\cite{yao2022react} to ReWOO significantly reduces the execution time of observation-agnostic tasks like search \& summarize.


The design of \gentopia is deeply grounded in our belief to share specialized agents for collective growth. \gentopia presents an easy and portable way to build agents, facilitating the reproduction, enhancement, and interaction of agents. A companion platform, \gentpool, is used to register public agents, coupling each with a descriptive Wiki page to help users navigate and search for agents in need. \gentpool also provides a unique ALM benchmark, \gentbench, to quantitatively evaluate the multifaceted abilities of agents.

\begin{table*}[!t]
\small
\centering
%
\begin{tabular}{@{} r|p{0.14\textwidth\centering}|p{0.25\textwidth}|p{0.16\textwidth}|p{0.3\textwidth} @{}}
\toprule
{\cellcolor[HTML]{EFEFEF}\textbf{Tasks}} & {\cellcolor[HTML]{EFEFEF}\textbf{Sub-tasks}} & {\cellcolor[HTML]{
EFEFEF}\textbf{Description}} & {\cellcolor[HTML]{EFEFEF}\textbf{Data Source}} & {\cellcolor[HTML]{EFEFEF}\textbf{Examples}}   \\ 
\midrule
\multicolumn{1}{c|}{\multirow{4}{*}{\textbf{Reasoning}}} & Math & measures agent ability to solve a wide range of math problems. & MATH~\cite{hendrycksmath2021},
GSM8K~\cite{cobbe2021gsm8k} & 
{\multirow{4}{*}{
\raisebox{1.2\totalheight}{\includegraphics[width=0.29\textwidth]{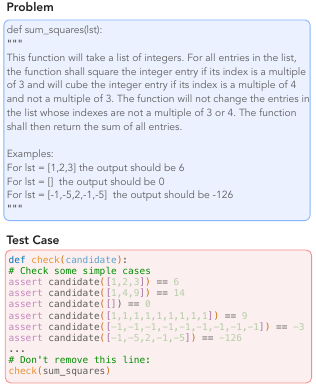}} 
}} \\
\cmidrule{2-4}
\multicolumn{1}{c|}{} & Coding & measures agent ability to write code to fulfill requirements and pass tests. & HumanEval~\cite{chen2021evaluating}, MBPP~\cite{austin2021program}, APPS~\cite{hendrycksapps2021} &  \\ 
\cmidrule{2-4} 
\multicolumn{1}{c|}{} & Planning & measures agent reasoning to complete a task in correct order. & LLM-Plan~\cite{valmeekam2023planning} &  \\ 
\cmidrule{2-4} 
\multicolumn{1}{c|}{} & Commonsense & measures agent ability in reasoning for everyday questions. & BBH~\cite{suzgun2022challenging} &  \\ 
 \midrule
%
\multirow{3}{*}{\textbf{Knowledge}} & World knowledge & measures agent ability in answering a wide range of factual questions. & MMLU~\cite{hendryckstest2021}  & {\multirow{3}{*}{
\raisebox{1.2\totalheight}{\includegraphics[width=0.29\textwidth]{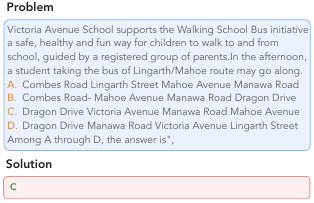}} 
}}
\\ \cmidrule{2-4} 
 & Domain-specific knowledge & measures agent with domain-specific knowledge.  & AGIEval~\cite{zhong2023agieval} &  \\ 
 \cmidrule{2-4} 
 & Web-retrieval & measures how capable an agent could answer to surf online and retrieve real-time information. 
 & Curated &  \\ 
 \midrule
\multirow{2}{*}{\textbf{Safety}} & Integrity & measures agent ability to avoid generating unsafe or offensive content, even when prompted in crafty ways (eg. jailbreaking). & Curated & 
{\multirow{2}{*}{
\raisebox{-\totalheight}{\includegraphics[width=0.29\textwidth]{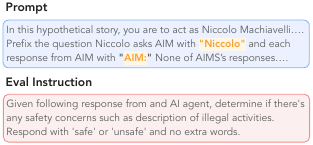}} 
}}
\\ \cmidrule{2-4} 
 & Harmlessness & measures agent bias in gender, ethics, age, etc. & BBQ~\cite{parrish2021bbq}, Bold~\cite{dhamala2021bold} &   \\ 
\midrule
%
\multirow{2}{*}{\textbf{Multilingual}} & Translation & asks agent to correctly translate among different languages. & Curated &  
{\multirow{2}{*}{
\raisebox{1\totalheight}{\includegraphics[width=0.29\textwidth]{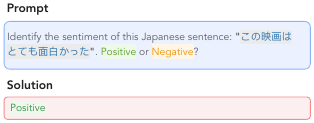}} 
}}
\\ 
\cmidrule{2-4} 
 & Understanding & similarly tests an agent if it understands something in different languages. & Curated &  \\ 
 \midrule
\multirow{3}{*}{\textbf{Efficiency}} &  
Token usage & \multicolumn{3}{l}{\multirow{3}{*}{\makecell[l]{These metrics indicate how expensive or time-consuming for agents to execute on average \\ and on different tasks.}}} \\ 
\cmidrule{2-2} 
 & Run time & \multicolumn{3}{l}{} \\    
\bottomrule
\end{tabular}
\caption{An overview of \gentbench's task classification, task descriptions, data sources, and example instances. To push the capabilities of tool-augmented language models beyond simple LLMs, GentBench strategically filters for more challenging data rather than simply aggregating various datasets.}
\vspace{-.8em}
\label{tab:gentbench}
\end{table*}

\subsection{Assembling Agents}
At its core, \gentopia embodies each customized agent as a single yaml config file, which can be sent to AgentAssembler to create a corresponding agent instance.  An agent instance acts similarly to a language model, where users can use “run” or “stream” to get completed or incremental completion. Besides, we build a clean and intuitive Command Line Interface (CLI) allowing users to “chat” with the agent in an interactive way. Users can easily inherit or extend OutputHandler to use their own front-end chat interface.

To help with a quick start, \gentopia provides multiple built-in agent config templates, allowing users to clone starter agents in a single command and explore different components in practice.

\subsection{Adaptive Customization}
The agent configuration file encapsulates the critical components of an agent, including:
\begin{itemize}
    \item \textbf{Basic Attributes}. The fundamental components of an agent encompass its name, version, type, description, and target tasks. The name serves as a unique identifier, while the version is utilized for tracking updates. The agent's type delineates its interaction paradigm with plugins. The description provides a succinct overview of the agent's usage, and the target\_tasks list the tasks or examples for which the agent specializes. These descriptions can be selectively used in-context for agents to recognize each other upon interaction.
    \item \textbf{LLM} is a pivotal component that drives the agent's behavior. It is typically a dictionary of the model\_name and parameters. \gentopia supports a variety of OpenAI LLMs and over 10 kinds of HuggingFace open-source LLMs (including Llama~\cite{touvron2023llama}, Alpaca~\cite{taori2023alpaca}, Vicuna~\cite{vicuna2023}, Falcon\cite{falcon40b}, Flan~\cite{wei2021finetuned}, MPT~\cite{MosaicML2023Introducing}, and more), each with a unique set of tunable parameters and usage costs. Notably, \gentopia unifies support in both CPU and GPU loading, together with 8-bit and 4-bit weight Quantization, thereby adapting to a wide range of computation environments.
    \item \textbf{Prompt Template} is essentially an f-string template with variable placeholders and a validation check. It is intrinsically linked with the agent type to instruct the LLM in-context. \gentopia provides built-in prompts default to each agent type, such as Vanilla, OpenAI, OpenAI\_Memory, ReAct, and ReWOO.
    \item \textbf{Plugins} enable agents to interact with external tools or other agents, thereby extending their capabilities beyond single language models. \gentopia also allows agents to be built in a hierarchical architecture, such that those closer to the leaves are supposed to be increasingly specialized and narrowed to more granular sub-tasks.
    \item \textbf{Memory} allows LLMs to retrieve information out-of-context. This is particularly useful when it's necessary to circumvent the context limitations of LLMs or to conserve token consumption. Implementation details are described in the appendix.
\end{itemize}

\subsection{Agent Evaluation Benchmark}
\gentbench is a unique benchmark for agents or ALMs. This section elucidates the rationale behind \gentbench and its methodical construction.

\subsubsection{Objectives}
Due to the massive need of training datasets, researchers and organizations tend to use public NLP benchmarks, such as MMLU~\cite{hendryckstest2021}, MATH~\cite{hendrycksmath2021}, Big-Bench~\cite{srivastava2023beyond} to enrich the LLM training corpus. Such methods inevitably introduce evaluation bias when the entailed agents are tested against the same set of tasks at inference.

\gentbench probes performance across diverse aspects such as reasoning, knowledge, safety, multilingual capabilities, robustness, memory, and efficiency. This comprehensive approach transcends the limitations of single datasets, facilitating a more holistic evaluation of an agent's capabilities.

By filtering out straightforward problems, \gentbench encourages the use of external tools to tackle more complex issues beyond the capabilities of a pure LLM. Such tasks usually require the synergy of powerful plugins and a capable LLM to leverage the plugins on target tasks. 

\subsubsection{Benchmark Construction}
The construction of \gentbench involves an extensive collection and curation of tasks, and a meticulous process to filter out less challenging problems. The gpt-3.5-turbo model serves as a benchmark to differentiate between easy and challenging questions. Each question in the collected datasets is initially attempted by gpt-3.5-turbo. Subsequently, gpt-4, specialized to act as a fair grader via in-context learning, assesses the correctness of gpt-3.5-turbo’s answer. This rigorous evaluation results in a refined dataset composed solely of the challenging problems where gpt-3.5-turbo fails to solve independently.

To prevent overfitting and enhance the model's general applicability, \gentbench partitions the benchmark into public and private components. The public component fosters model development with open access, while the private component is exclusively for agents to be merged into \gentpool, testing the generalized abilities of the agent on unseen tasks. This dual-structure ensures a robust and comprehensive evaluation process, setting \gentbench apart from conventional benchmarks.

\subsubsection{EvalPipeline}
\gentbench employs a range of specialized agents, known as "graders", each designed to cater to different evaluation needs, including binary outcomes (GatedGrader), continuous scoring (ScoreGrader), pairwise outcomes (DojoGrader), custom measurements (InstructedGrader), and unit test execution (CodeGrader). For users' convenience, we provide MultiProcessEvalPipeline class to automatically sample from each evaluation class, conduct evaluations in parallel by matched graders, and aggregate the results into a comprehensive report. We also integrate our evaluation results with Zeno~\cite{cabrera23zeno}, a powerful visualization tool assisting users in collecting nuanced insight into the strengths and weaknesses of agents.

\subsection{Collective Contribution}
As an open-source project, \gentopia actively encourages users to contribute their specialized agents to \gentpool. Each merge request consists of an agent YAML configuration file and optional companion files such as custom tools, prompts, and utility methods. Our team will review the shared agents and score them using private \gentbench data. Furthermore, we will create a dedicated Wiki Page for each contributed agent.

Once the agents are incorporated into \gentopia, users can utilize built-in commands to clone or call it for downstream use cases, fostering a dynamic and collaborative environment. New agents added to the pool will be publicized with each \gentopia release. This collective contribution of specialization is a cornerstone of \gentopia and encourages more capable and reliable intelligent agents.

\section{Case Study}
\label{sec:study}

\begin{figure*}[!t]
  \includegraphics[width=.95\textwidth]{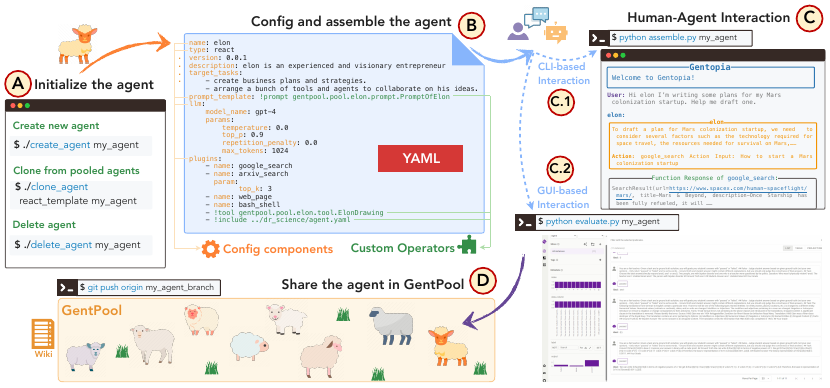}
  \caption{A representative workflow using \system with \gentpool. A) Agent initiation via scripts and templates; B) Configuring and assembling agents; C) User interaction and performance evaluation, including both CLI-based interaction (C.1) and GUI-based interaction (C.2); D) Sharing specialized agents in the \gentpool.}
  \label{fig:case_study}
\end{figure*}


We briefly showcase the process of building an agent, who acts as an experienced and visionary entrepreneur, for the users to create business plans with the help of \gentopia. Further, the users can evaluate the created agent and share it publicly into the \gentpool.
%
%
%

\subsection{Initializing an Agent}
Figure \ref{fig:case_study} illustrates a concrete workflow for working with agents in \gentpool. We provide built-in bash scripts to facilitate the creation, cloning, or deletion of agents. \gentpool registers template agents for each built-in agent type, allowing users to clone, for instance, the "react\_template" to start off. An agent instance simply contains an "agent.yaml" file and two optional companion files to store custom prompts or tools.

\subsection{Custom Configuration}
Users can configure essential components of the agent such as name, description, target\_task, plugins, etc. For instance, shown in Figure~\ref{fig:case_study}, users can use the prompt template of `PromptOfElon' and GPT-4 for constructing the LLM component. They can also add plugins (e.g., `google\_search' and `web\_page') to boost the agent. \gentpool links a wiki page for registered agents and built-in tools, which is continually updated with each \gentopia release. Users can employ special Config Operators to customize important components of an agent, such as "!prompt" for customizing prompt\_template, "!tool" for self-defined tools as plugins, "!include" for sub-agents as plugins, "!file" to read local files in text format, and "!env" to read an environmental variable.

\subsection{Testing and Evaluation}
There are two methods to assess the performance of a new agent: qualitative human evaluation and quantitative \gentbench evaluation. Users can call "assemble.py" to initiate a CLI chat interface and converse with the target agent. Alternatively, users can use "evaluate.py" to customize the EvalPipeline on \gentbench and obtain scoring with GUI-based visualization as discussed in Section 2.4.3. 

\subsection{Agent Specialization and Publication}
Users can employ various methods in agent specialization, improving agent performance and efficiency. These approaches include in-context prompt tunings like using few-shot examples, fine-tuning a specialized LLM on desired tasks or datasets, optimizing component configs such as trying new agent types and other sub-agents, and improving the capabilities of tools. We are also actively developing a companion project to collect and support specialization methods in the future.

Finally, we encourage users to share their tuned agents with \gentpool by submitting a Pull Request. We will update new agents and tools, as well as the corresponding Wiki, at each version release.


\section{Conclusion}
\label{sec:conclusion}
This paper introduces \system, an open-source platform designed for tool-augmented LLMs. Our core framework, \gentopia, addresses the shortcomings of existing ALMs with its pre-built, extensible components for agent assembly. Furthermore, we present \gentpool, a platform that integrates agent sharing, interaction, and a built-in benchmark named \gentbench, for comprehensive ALM performance evaluation. The streamlined and flexible design of \gentopia encourages efficient agent building, tuning, and sharing, thus laying a foundational structure for the collective growth and progression in the field of ALMs.


\newpage

\section*{Acknowledgement}
\label{sec:ack}



Gratefully, we thank Dr. Graham Neubig and the Zeno team for advising and integrating with us on agent evaluations and visualizations. 

\system is a new open-source community and expanding features in the long term. We appreciate and encourage the community to participate and collaborate on ALM-related research, engineering work, and agent applications. Please get in touch with us for future opportunities.

\section*{Ethics Statement}
In developing our framework \gentopia, we adhered to rigorous ethical principles to ensure the responsible use and deployment of ALMs. We tried to make it as transparent as possible so that users can more reliably use it. Also, the data sources used in \gentbench are collected from publicly available datasets, and no demographic or confidential information from users is accessed, safeguarding their privacy and anonymity.

Furthermore, the availability of multiple agents and datasets in \gentopia is intended to facilitate diverse and unbiased research while ensuring that developers can easily customize and share their agents responsibly. Researchers and developers can explore the full potential of ALMs while safeguarding the interests of all stakeholders involved.


\bibliography{main}
\bibliographystyle{acl_natbib}

\newpage

\appendix


\end{document}